%% file: main.tex
\documentclass[10pt,twocolumn,letterpaper]{article}

\usepackage[pagenumbers]{cvpr} 

\input{preamble}

\definecolor{cvprblue}{rgb}{0.21,0.49,0.74}
\usepackage[pagebackref,breaklinks,colorlinks,allcolors=cvprblue]{hyperref}
\usepackage{bm}
\usepackage{amsmath}
\usepackage{makecell}
\usepackage{multirow}

\def\paperID{2822} 
\def\confName{CVPR}
\def\confYear{2025}

\title{Generative Image Layer Decomposition with Visual Effects}

\author{
Jinrui Yang$^{1,2,*}$\hspace{0.2cm}
Qing Liu$^{2}$ \hspace{0.2cm}
Yijun Li$^{2}$ \hspace{0.2cm}
Soo Ye Kim$^{2}$ \hspace{0.2cm}
Daniil Pakhomov$^{2}$ \\
Mengwei Ren$^{2}$\hspace{0.2cm}
Jianming Zhang$^{2}$ \hspace{0.2cm}
Zhe Lin$^{2}$\hspace{0.2cm}
Cihang Xie$^{1}$\hspace{0.2cm}
Yuyin Zhou$^{1}$ \\
$^1$UC Santa Cruz
\hspace{0.8cm}
$^2$Adobe Research \hspace{0.8cm} \\
\url{https://rayjryang.github.io/LayerDecomp}
\vspace{-0.4cm}
}

\begin{document}

\twocolumn[{%
\renewcommand\twocolumn[1][]{#1}%
\maketitle
\begin{center}
    \centering
    \captionsetup{type=figure}
\includegraphics[width=1.0\linewidth]{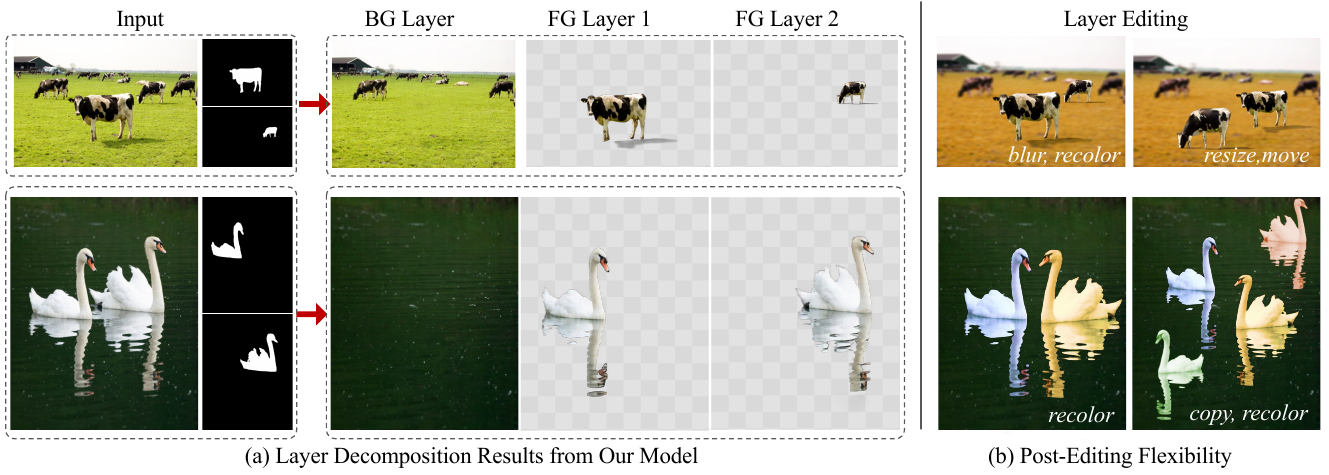}
    \captionof{figure}{
   (a) Given an input image and a binary object mask, our model is able to decompose the image into a clean background layer and a transparent foreground layer with preserved visual effects such as shadows and reflections. (b) Subsequently, our decomposition empowers complex and controllable layer-wise editing such as spatial, color and/or style editing.} 
    \label{fig:teaser_image}
\end{center}

}]

\maketitle

\input{sec/0_abstract}

\input{sec/1_intro}
\input{sec/2_related_work}
\input{sec/3_approach}
\input{sec/4_experiment}
\input{sec/5_conclusion}

{
    \small
\bibliographystyle{ieeenat_fullname}
    \bibliography{main}
}

\input{sec/X_suppl}


\end{document}

%% file: preamble.tex
\usepackage[symbol]{footmisc}

%% file: sec/0_abstract.tex
\begin{abstract}

Recent advancements in large generative models, particularly diffusion-based methods, have significantly enhanced the capabilities of image editing. However, achieving precise control over image composition tasks remains a challenge. 
Layered representations, which allow for independent editing of image components, are essential for user-driven content creation, yet existing approaches often struggle to decompose image into plausible layers with accurately retained transparent visual effects such as shadows and reflections. We propose \textbf{LayerDecomp}, a generative framework for image layer decomposition which outputs photorealistic clean backgrounds and high-quality transparent foregrounds with faithfully preserved visual effects.
To enable effective training, we first introduce a dataset preparation pipeline that automatically scales up simulated multi-layer data with synthesized visual effects. To further enhance real-world applicability, we supplement this simulated dataset with camera-captured images containing natural visual effects. 
Additionally, we propose a consistency loss which enforces the model to learn accurate representations for the transparent foreground layer when ground-truth annotations are not available.
Our method achieves superior quality in layer decomposition, outperforming existing approaches in object removal and spatial editing tasks across several benchmarks and multiple user studies, unlocking various creative possibilities for layer-wise image editing. 
The project page is \url{https://rayjryang.github.io/LayerDecomp}.
\footnotetext{*This work was done when Jinrui Yang was a research intern at Adobe Research.}
\end{abstract}

%% file: sec/1_intro.tex
\vspace{-2em}
\section{Introduction}
\label{sec:intro}

The rapid advancement of large-scale text-to-image diffusion models~\cite{rombach2022high,esser2024scaling,betker2023improving} has greatly improved image editing capabilities, with recent studies~\cite{jia2024designedit,sheynin2024emu,zhao2024ultraedit,brooks2023instructpix2pix} demonstrating promising results by training on large-scale datasets of captioned images. However, achieving precise control for user-driven image composition tasks remains challenging. 
Layered representations, which decompose image components into independently editable layers, are essential for precise user-driven content creation. Most visual content editing
software and workflows are layer-based, relying heavily on transparent or layered elements to compose and create content. Despite this, few existing approaches have explored layer-based representation for image editing in depth. Recently, LayerDiffusion~\cite{zhang2024transparent} is proposed to generate transparent layer representations from text inputs; however, this approach is not well-suited for image-to-image editing tasks. Meanwhile, MULAN~\cite{tudosiu2024mulan} presents a multi-layer annotated dataset for controllable generation, but it often fails to preserve essential visual effects in the right layers, limiting its adaptability for seamless downstream edits.

This work aims to fill these gaps by decomposing an input image into two layers: a highly plausible clean background layer, and a high-quality transparent foreground layer that retains natural visual effects associated with the target. These decomposed layers will support layer-constrained content modification and allow seamless blending for harmonious re-composition. Additionally, the faithfully preserved natural visual effects in the foreground layer will benefit related research areas, such as shadow detection and shadow generation.
However, the lack of publicly available multi-layer datasets with realistic visual effects, such as shadows and reflections, poses a significant challenge for training high-quality layer-wise decomposition models. How to accurately decompose images into layers and learn correct representation for visual effects in the foreground layer without ground-truth data is the key problem to solve.

To address these challenges, we first design a dataset preparation pipeline to collect data from two sources: (1) simulated image triplets consisting of composite images blended from random background and unrelated transparent foreground with generated shadow, allowing us to create a large-scale training set with ground-truth for both branches; and (2) camera-captured images for a scene with and without the target foreground object, ensuring the model can effectively adapt to real-world scenarios.
Building on this dataset, we introduce \textbf{\underline{Layer} \underline{Decomp}osition with Visual Effects (\ours{})}, a generative training framework that enables large-scale pretrained diffusion transformers to effectively decompose images into editable layers with correct representation of visual effects. 
The key to our model training lies in a consistency loss, which ensures faithful retention of natural visual effects within the foreground layer while maintaining background coherence.
Specifically, for real-world data where visual effect annotations are not available, it is not feasible to directly compute diffusion loss on the foreground layer. Instead, \ours{} enforces consistency between the original input image and the re-composite result, blended from the two predicted layers, to ensure the model learns correct representation for the transparent foreground layer.

As shown in Figure~\ref{fig:teaser_image}, \ours{} can effectively decompose an input image into a clean background and a transparent foreground with preserved visual effects. This capability supports downstream editing applications without requiring additional model training or inference. Furthermore, our method exhibits superior layer decomposition quality, outperforming existing approaches in both object removal tasks and object spatial editing tasks across various benchmarks and in multiple user studies.

In summary, our contributions are as follows:

\begin{itemize}
    \item  We introduce a scalable pipeline to generate large-scale simulated multi-layer data with paired ground-truth visual effects for training layer decomposition models.  
    
    \item We present \ours, a generative training framework that leverages both simulated and real-world data to enable robust layer-wise decomposition with accurate visual effect representation through a consistency loss, facilitating high-quality, training-free downstream edits.
   
    \item \ours~surpasses existing state-of-the-arts in maintaining visual integrity during layer decomposition, excelling in object removal and object spatial editing, and enabling more creative layer-wise editing.
    
\end{itemize}


\begin{figure*}[t]
  \centering
   \includegraphics[width=1.0\linewidth]{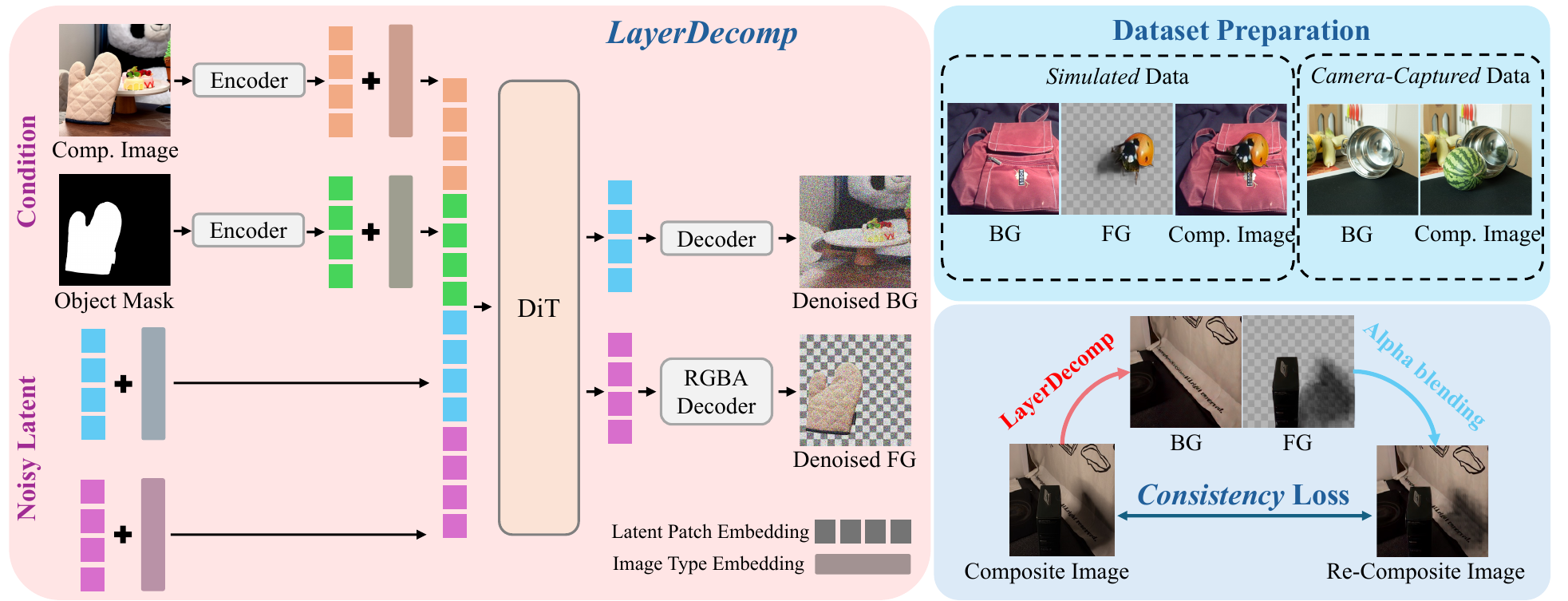}
   \caption{
   \textbf{The framework of \ours{}}. The model takes four inputs: two conditional inputs, including a composite image and an object mask, and two noisy latent representations of the background and foreground layers. During training, we use simulated image triplets alongside camera-captured background-composite image pairs. We also introduce a pixel-space consistency loss to ensure that natural visual effects such as shadows and refelctions are faithfully preserved in the transparent foreground layer.}
   \label{fig:framework}
\end{figure*}

%% file: sec/2_related_work.tex
\section{ Related Works}
\label{sec:related_work}

\subsection{ Image Editing }

Image editing methods can be broadly categorized into two groups: multi-task editing and local editing. Multi-task editing allows for a wide array of image modifications based on high-level inputs, such as user instructions. For example, Emu-Edit~\cite{sheynin2024emu} offers a flexible interface for diverse image edits. Other methods, including InstructPix2Pix~\cite{brooks2022instructpix2pix}, Prompt2Prompt~\cite{hertz2022prompt}, InstructAny2Pix~\cite{li2023instructany2pix}, MagicBrush~\cite{zhang2024magicbrush}, Imagic~\cite{kawar2023imagic}, PhotoSwap~\cite{gu2024photoswap}, UltraEdit~\cite{zhao2024ultraedit}, HQ-Edit~\cite{hui2024hq}, MGIE~\cite{fu2024mgie}, and OmniGen~\cite{xiao2024omnigen}, focus on fine-grained, user-specific edits. These approaches achieve high precision by aligning closely with user instructions, often through fine-tuned models. However, due to the absence of region references, these methods struggle with accurately locating objects and preserving the integrity of unrelated regions.

Local image editing focuses on tasks like inpainting and object insertion, typically guided by masks or reference images. Early GAN-based methods, such as CMGAN~\cite{zheng2022image}, LAMA~\cite{suvorov2022resolution}, CoModGAN~\cite{zhao2021large}, ProFill~\cite{zeng2020high}, CRFill~\cite{zeng2021cr}, and DeepFillv2~\cite{yu2019free}, use latent space manipulation for inpainting missing regions.
Leveraging the success of text-to-image diffusion models,  Repaint~\cite{lugmayr2022repaint}, SDEdit~\cite{meng2021sdedit}, ControlNet~\cite{zhang2023adding}, BrushNet~\cite{ju2024brushnet}, and Blended Diffusion~\cite{avrahami2022blended} combine text guidance with masks or references for customized image inpainting. 
Further, ObjectStitch~\cite{song2023objectstitch}, HDPainter~\cite{manukyan2023hd}, and PowerPaint~\cite{zhuang2023task} extend this to versatile object editing with text prompts.
Beyond local object editing, recent research also focus on spatial editing techniques that allow more interactive control over object positioning and transformations. Methods like MagicFixup~\cite{alzayer2024magic}, DiffusionHandle~\cite{pandey2024diffusion}, DragGAN~\cite{pan2023drag}, DiffEditor~\cite{mou2024diffeditor}, DesignEdit~\cite{jia2024designedit}, and DragAnything~\cite{wu2025draganything} highlight a growing emphasis on user-driven, spatially aware editing. However, preserving object appearance and background integrity during spatial edits (e.g., moving and resizing) remains challenging, particularly when complex visual effects like shadows and reflections are involved.

\subsection{ Image Layer representation }

Obtaining high-quality image layer representations is crucial for implementing accurate and diverse editing objectives. This process typically involves image decomposition, layer extraction, and image matting. PACO~\cite{liu2024object} provides a fine-grained dataset with mask annotations for parts and attributes of common objects. However, these object representations indicate only the regions of parts and objects, lacking transparent layers for flexible editing.
MAGICK~\cite{burgert2024magick} offers a large-scale matting dataset generated by diffusion models, while MULAN~\cite{tudosiu2024mulan} creates RGBA layers from COCO~\cite{lin2014microsoft} and Laion Aesthetics 6.5~\cite{schuhmann2022laion} using off-the-shelf detection, segmentation, and inpainting models; however, it does not capture visual effects, limiting its applicability for direct editing. 
LayerDiffusion~\cite{zhang2024transparent} and Alfie~\cite{quattrini2024alfie} generate transparent image layers from text prompts, facilitating layer blending. However, text-driven generation limits object identity control in image-specific editing tasks.
Once the image decomposition is performed, users typically need to execute image composition to achieve a cohesive final image. Existing methods like ObjectDrop~\cite{winter2024objectdrop} require model fine-tuning to restore visual effects, which can disrupt the original image’s appearance. In contrast, \ours~ models these effects during training, inherently preserving them to produce a seamless, harmonious composite image without extra fine-tuning.

%% file: sec/3_approach.tex
\section{Approach}
\label{sec:approach}

\subsection{Overview of \ours{} Framework}
As shown in Fig~\ref{fig:framework}, the \ours{} framework builds upon Diffusion Transformers (DiTs)~\cite{peebles2023scalable} to denoise multi-layer image outputs in the latent space encoded by the VAE encoders $g_\phi^{\text{RGB}}(\cdot)$ and $g_\psi^{\text{RGBA}}(\cdot)$. Specifically, the DiT model $f_\theta(\cdot)$ takes two types of conditional input $\mathbf{c} \!=\! (\mathbf{y}_\text{comp}, \mathbf{y}_\text{obj})$, which are the latents of the original composite image,  and the decomposition object mask, respectively, \textit{i.e.}, $\mathbf{c} \!\!\!=\!\!\! (\mathbf{y}_\text{comp}, \mathbf{y}_\text{obj}) \!\!=\!\! (g_\phi^{\text{RGB}}(\mathbf{I}^{\text{RGB}}_{\text{comp}}), g_\phi^{\text{RGB}}(\mathbf{M}_{\text{obj}})) $. By taking the conditional image embeddings, the model targets to denoise the noisy latent $\mathbf{x}_t \!\!=\!\! (\mathbf{x}_t^{\text{bg}}, \mathbf{x}_t^{\text{fg}})$, to recover the latents of the clean background image and the transparent foreground layer $ \mathbf{x}_0 \!\!=\!\! (\mathbf{x}_0^{\text{bg}}, \mathbf{x}_0^{\text{fg}}) \!=\! (g_\phi^{\text{RGB}}(\mathbf{I}^{\text{RGB}}_{\text{bg}}), g_\psi^{\text{RGBA}}(\mathbf{I}^{\text{RGBA}}_{\text{fg}}))$. The training loss follows the standard denoising diffusion loss~\cite{ho2020denoising,lipman2022flow}: 
\begin{align}
    &\mathcal{L}_{\mathrm{dm}} =  \mathbb{E}_{t\sim \mathcal{U}(\{1,...,T\}), \boldsymbol{\epsilon}, \mathbf{x}_t } \left[ \| \boldsymbol{\epsilon}_\theta(\mathbf{x}_t; \mathbf{c}, t) - \boldsymbol{\epsilon} \|_2^2\right], \label{eq:dm}\\
    &s.t.~~ \mathbf{x}_0 \!\sim\! q_{\text{data}}(\mathbf{x}_0), \boldsymbol{\epsilon} \!\sim\! \mathcal{N}(\boldsymbol{0}, I), \mathbf{x}_t = \sqrt{\alpha_t} \mathbf{x}_0 + \sqrt{1 -\alpha_t} \boldsymbol{\epsilon} \notag.
\end{align}

Specifically, the noisy input $\mathbf{x}_t$ and image conditions $\mathbf{c}$ are initially divided into non-overlapping patches and converted into patch embeddings. The patch embeddings of each type of images, such as background, foreground or any conditions, are added with a corresponding \textit{type embedding} and then concatenated into a sequence. Subsequently, the model follows the standard DiT architecture, where the patch embeddings are processed through multiple transformer blocks. With such design, the image conditions provide comprehensive contextual information through the self-attention in the transformer blocks to enhance the denoising, and the loss is only computed on the positions corresponding to the noisy latents. 

Note that the latent of the foreground image needs to be encoded with RGBA channels, we leverage an RGBA-VAE fine-tuned from the original VAE by following LayerDiffusion~\cite{zhang2024transparent} which makes minimal disturbance to the original latent space. Moreover, as the foreground image is not always available, \textit{e.g.}, in the case of real-world camera-captured data, the noisy input and output corresponding to the foreground are masked out from $\mathcal{L}_{\mathrm{dm}}$ computation if $\mathbf{I}^{\text{RGBA}}_{\text{fg}}$ is absent in the training stage.

\subsection{Consistency Loss for Visual Effects Learning}
To handle cases where real-world data lacks ground-truth annotations, we introduce a consistency loss that enables the learning of natural visual effects in the transparent foreground layer without explicit annotation. Intuitively, as shown in Fig~\ref{fig:framework}, the consistency loss is applied in the decoded pixel space to encourage the predicted foreground can faithfully reconstruct the composite input after blending with the background layers.

More specifically, given a composite image 
$\mathbf{{I}}^{\text{RGB}}_{\text{comp}}$, at any denoising timestep $t$, we reparameterize our model prediction back to the estimation of the clean latent $\mathbf{x}_0$ as: 
\begin{equation}
\hat{\mathbf{x}}_0(\mathbf{x}_t; \mathbf{c}, t) = \frac{1}{\sqrt{\alpha_t}} \left(\mathbf{x}_t - \sqrt{1 - \alpha_t} \cdot \boldsymbol{\epsilon}_\theta(\mathbf{x}_t; \mathbf{c}, t)\right).
\label{eq:recovered_x0}
\end{equation}

Given Eq.~\ref{eq:recovered_x0}, we compute the estimated $\hat{\mathbf{x}}_0(\mathbf{x}_t; \mathbf{c}, t) = (\hat{\mathbf{x}}_0^{\text{bg}}, \hat{\mathbf{x}}_0^{\text{fg}})$ of background and foreground at time step $t$ and decode them into pixel space to get $\mathbf{\hat{I}}^{\text{RGB}}_{\text{bg}} = h_{\phi^\prime}^{\text{RGB}}(\hat{\mathbf{x}}_0^{\text{bg}})$ and $\mathbf{\hat{I}}^{\text{RGBA}}_{\text{fg}}=h_{\psi^\prime}^{\text{RGBA}}(\hat{\mathbf{x}}_0^{\text{fg}})$, via the decoder of RGB-VAE and RGBA-VAE, respectively. The results are combined through alpha blending to produce the estimated composite image $\mathbf{\hat{I}}^{\text{RGB}}_{\text{comp}} = \mathcal{A}(\mathbf{\hat{I}}^{\text{RGB}}_{\text{bg}}, \mathbf{\hat{I}}^{\text{RGBA}}_{\text{fg}})$. The consistency loss is thus:
\begin{equation}
\mathcal{L}_{\text{consist}} = \mathbb{E}_t \sum_{i=1}^H \sum_{j=1}^W \left| \mathbf{{I}}^{\text{RGB}}_{\text{comp}}(i,j) - \mathbf{\hat{I}}^{\text{RGB}}_{\text{comp}}(i,j) \right|,
\end{equation}
where $H$ and $W$ indicates the height and the width of the composite image, respectively.

The consistency loss enables \ours{} to learn faithful representations of transparent visual effects in the foreground layer, which is essential for accurately decomposing natural shadows and reflections in real-world data, especially in the absence of ground-truth annotations.

\subsection{Dataset Preparation}
To effectively train \ours{}, we curated a hybrid dataset that combines simulated and real-world data. Ideally, training \ours{} requires image triplets: an input image $\mathbf{I}^{\text{RGB}}_{\text{comp}}$, a transparent foreground layer containing the target object and its visual effects $\mathbf{I}^{\text{RGBA}}_{\text{fg}}$, and a background image without the foreground object $\mathbf{I}^{\text{RGB}}_{\text{bg}}$. While collecting natural triplet images with specialized devices or through manual annotation might be feasible, it is costly and impractical for large-scale data needs. Conversely, synthesizing such triplet data directly with generative models presents significant challenges. Observations from existing approaches, such as HQ-Edit~\cite{hui2024hq} and LayerDiffusion~\cite{zhang2024transparent}, indicate that generative models often inadvertently modify areas outside the target foreground, making it difficult to produce truly aligned image layers with consistent content. Additionally, accurately representing transparent visual effects, such as shadows and reflections in the foreground layer with an alpha channel for transparency, remains unexplored in existing works. To address these limitations, we developed a simulated data pipeline to create triplet images and supplemented it with a smaller portion of real-world $\mathbf{I}^{\text{RGB}}_{\text{comp}}$ and $\mathbf{I}^{\text{RGB}}_{\text{bg}}$ pairs to enhance robustness.

\vspace{0.5em}
\noindent\textbf{Simulated Data}: To create image triplets, we first collected a large-scale object assets consisting of unoccluded foreground objects with synthesized shadows. We used entity segmentation~\cite{qi2022open} to select ``thing'' objects from natural images, and applied depth estimation to infer occlusion relations to exclude incomplete objects. We then applied a shadow synthesis method~\cite{metashadow} to generate a shadow intensity map for each object on a white background. By integrating the intensity map into the alpha channel, we obtained comprehensive object assets in RGBA format. During training, we adjusted the scale and position of each foreground asset to align with the properties of a randomly selected background image $\mathbf{I}^{\text{RGB}}_{\text{bg}}$, resulting in a finalized foreground layer $\mathbf{I}^{\text{RGBA}}_{\text{fg}}$. By blending the two layers together, we obtained a composite image $\mathbf{I}^{\text{RGB}}_{\text{comp}}$, completing the triplet data needed for model training. Although the composite results may lack fully realistic geometry and harmonized content, this approach enables large-scale training and allows the model to learn the appropriate representations for the two output layers in the decomposition task.

\vspace{0.5em}
\noindent\textbf{Camera-Captured Data}: We also include a small set of real-world camera-captured image pairs, denoted as $\mathbf{I}^{\text{RGB}}_{\text{com}}$ and $\mathbf{I}^{\text{RGB}}_{\text{bg}}$, similar to the \textit{counterfactual} dataset proposed by ObjectDrop~\cite{winter2024objectdrop}. The real-world data enhances the model's ability to generalize to natural images containing authentic shadows and a broader range of visual effects, such as reflections, which are crucial for accurate foreground-background decomposition in real-world tasks.

%% file: sec/4_experiment.tex
\section{Experiments}
\label{sec:experiment}

\begin{figure*}[t]
  \centering
   \includegraphics[width=1.0\linewidth]{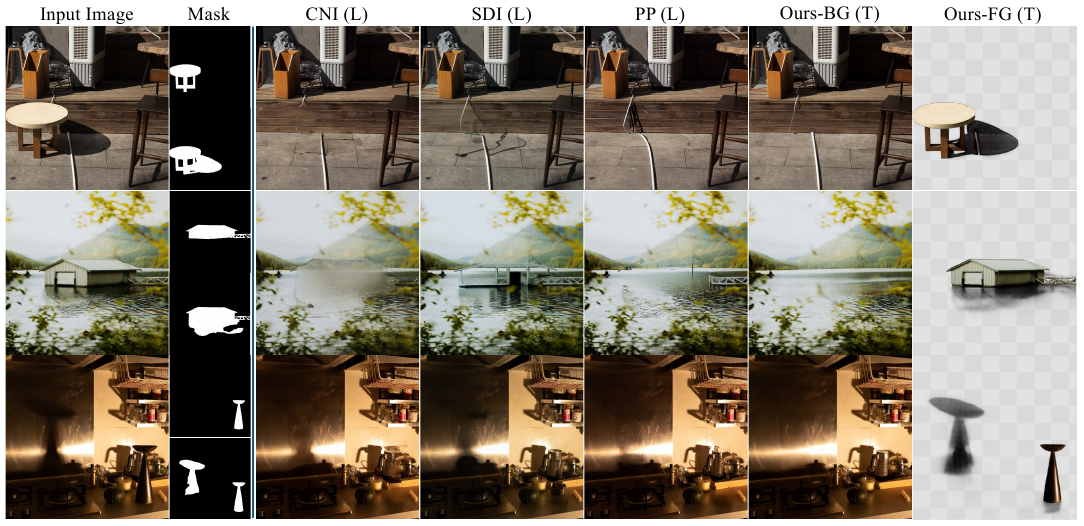}
   \caption{ \textbf{Object removal - comparison with mask-based methods.} Our model, using tight input masks, generates more visually plausible results with fewer artifacts compared to ControlNet Inpainting~\cite{zhang2023adding}, SD-XL Inpainting~\cite{rombach2022high}, and PowerPaint~\cite{zhuang2023task}, which all require loose mask input. Besides, our model delivers coherent foreground layers and supports more advanced downstream editing tasks.}
   \label{fig:comp_object_removal_mask_methods}
\end{figure*}

\textbf{Implementation.} \ours{} is finetuned from a 5 billion-parameter DiT model pre-trained for text-to-image generation. For layer decomposition task, the text encoder is dropped and the model does not take text input. RGBA-VAE is finetuned from the DiT VAE using a combination of L1 loss, GAN loss, and perceptual loss~\cite{zhang2018unreasonable}. It takes images with $512\times512$ resolution and encode them into $64\times64$ latent feature maps. Image type embedding is learnable through linear layers. Our simulated dataset is built from a large corpus of stock images, and our camera-captured dataset consists of $6,000$ image pairs. During training, we use the Adam optimizer and set the learning rate at $1e\text{-}5$. Training is conducted with a total batch size of $128$ on $16$ A100 GPUs for $80,000$ iterations. During inference, all results are generated using DDIM sampling with $50$ steps.

\begin{table}[t]
    \setlength\tabcolsep{2pt}
    \renewcommand{\arraystretch}{1.0}
    \footnotesize
    \centering
    \caption{\textbf{Ablation study of \ours{} on held-out test set.} ``BG'' denotes the decomposed background, and ``Comp'' represents the re-composited image created by our two-layer results.}
    \resizebox{\columnwidth}{!}{%
    \begin{tabular}{lcccccccc}
        \toprule
        Model & \multicolumn{2}{c}{PSNR $\uparrow$} & \multicolumn{2}{c}{LPIPS $\downarrow$} & \multicolumn{2}{c}{FID $\downarrow$} & \multicolumn{2}{c}{CLIP-FID $\downarrow$} \\
        \cmidrule(lr){2-3} \cmidrule(lr){4-5} \cmidrule(lr){6-7} \cmidrule(lr){8-9}
        & BG & Comp & BG & Comp & BG & Comp & BG & Comp \\
        \midrule
        $V_0$:RGB-only & 28.21 & - & 0.0732 & - & 21.00 & - & 4.551 & - \\
        $V_1$:$V_0+$RGBA FG (obj.) & $28.28$ & $27.53$ & $0.0708$ & $0.0649$ & $18.48$ & $18.83$ & $2.487$ & $2.329$ \\
        $V_2$:$V_0+$RGBA FG (obj.+v.e.)    & $28.56$ & $28.66$ & $0.0691$ & $0.0612$ & $17.99$ & $16.87$ & $2.539$ & $2.172$ \\
        \cmidrule(lr){1-9}
        Ours:$V_2+\mathcal{L}_{\text{consist}}$      & $29.27$ & $30.53$ & $0.0618$ & $0.0494$ & $16.04$ & $12.75$ & $1.813$ & $1.564$ \\
        \bottomrule
    \end{tabular}%
    }
    \label{tab:quant:ablation_study}
\end{table}

\begin{table*}[t]
    \centering
    \setlength\tabcolsep{3pt}
    \renewcommand{\arraystretch}{1.0}
    \footnotesize
    \caption{\textbf{Comparison of \ours{} with mask-based object removal methods.} Loose (L) and Tight (T) mask-based results are shown where applicable. \textbf{PSNR$^m$} and \textbf{SSIM$^m$} are computed on the shadow mask region to assess the model's shadow removal ability.}
    \resizebox{\textwidth}{!}{%
    \begin{tabular}{lcccccccccc}
        \toprule
        & \multicolumn{4}{c}{\textbf{RORD}~\cite{sagong2022rord}} & \multicolumn{4}{c}{\textbf{MULAN}~\cite{tudosiu2024mulan}} & \multicolumn{2}{c}{\textbf{DESOBAv2}~\cite{liu2023desobav2}} \\
        \cmidrule(lr){2-5} \cmidrule(lr){6-9} \cmidrule(lr){10-11}
        \textbf{Model} & \textbf{PSNR $\uparrow$} & \textbf{LPIPS $\downarrow$} & \textbf{FID $\downarrow$} & \textbf{CLIP-FID $\downarrow$} & \textbf{PSNR $\uparrow$} & \textbf{LPIPS $\downarrow$} & \textbf{FID $\downarrow$} & \textbf{CLIP-FID $\downarrow$} & \textbf{PSNR$^m$ $\uparrow$} & \textbf{SSIM$^m$ $\uparrow$} \\
        \midrule
        CNI~\cite{zhang2023adding} & $20.45^{L}22.01^{T}$ & $0.235^{L}0.182^{T}$ & $50.40^{L}53.71^{T}$ & $8.853^{L}9.262^{T}$ & $17.79$ & $0.321$ & $65.03$ & $9.396$ & $36.94^{L}$ & $0.491^{L}$ \\
        SDI~\cite{rombach2022high} & $19.88^{L}20.81^{T}$ & $0.205^{L}0.166^{T}$ & $53.73^{L}56.28^{T}$ & $11.38^{L}11.10^{T}$ & $16.04$ & $0.303$ & $65.74$ & $11.54$ & $34.21^{L}$ & $0.527^{L}$ \\
        PP~\cite{zhuang2023task}  & $20.88^{L}21.26^{T}$ & $0.231^{L}0.201^{T}$ & $39.48^{L}56.56^{T}$ & $8.596^{L}11.32^{T}$ & $17.17$ & $0.314$ & $55.80$ & $9.988$ & $29.33^{L}$ & $0.369^{L}$ \\
        \cmidrule(lr){1-11}
        Ours & $\mathbf{24.56}^{L}\mathbf{24.79}^{T}$ & $\mathbf{0.133}^{L}\mathbf{0.132}^{T}$ & $\mathbf{21.77}^{L}\mathbf{21.73}^{T}$ & $\mathbf{5.735}^{L}\mathbf{5.778}^{T}$ & $\mathbf{19.13}$ & $\mathbf{0.244}$ & $\mathbf{39.26}$ & $\mathbf{6.332}$ & $\mathbf{38.57}^{T}$ & $\mathbf{0.640}^{T}$ \\
        \bottomrule
    \end{tabular}%
    }
    \label{tab:quant:remove}
\end{table*}

\subsection{Ablations}
To quantitatively assess the advantages of incorporating visual effects in the foreground layer and the effectiveness of our proposed consistency loss, we construct a held-out evaluation dataset of $635$ images for ablation studies. This dataset includes camera-captured composite images with the corresponding backgrounds. Visual examples will be provided in the supplementary materials. 
The quality of the decomposed background layers can be directly evaluated using standard metrics, such as PSNR, LPIPS~\cite{zhang2018unreasonable}, FID~\cite{heusel2017gans,parmar2021cleanfid}, and CLIP-FID~\cite{parmar2021cleanfid}. To further evaluate the quality of the decomposed foreground layers, we apply alpha blending to re-composite the background and foreground together, comparing the result with the original composite image. As shown in Tab.~\ref{tab:quant:ablation_study}, compared to a na\"{i}ve DiT baseline which outputs only an RGB background, adding an RGBA foreground layer not only enables decomposition but also improves background quality. Incorporating visual effects in the foreground layer and introducing a consistency loss further enhance model performance. This demonstrates the decomposition task may implicitly improve the model's understanding of the input scene, leading to superior results in both layers' predictions.

\begin{figure}[t]
  \centering
   \includegraphics[width=1.0\linewidth]{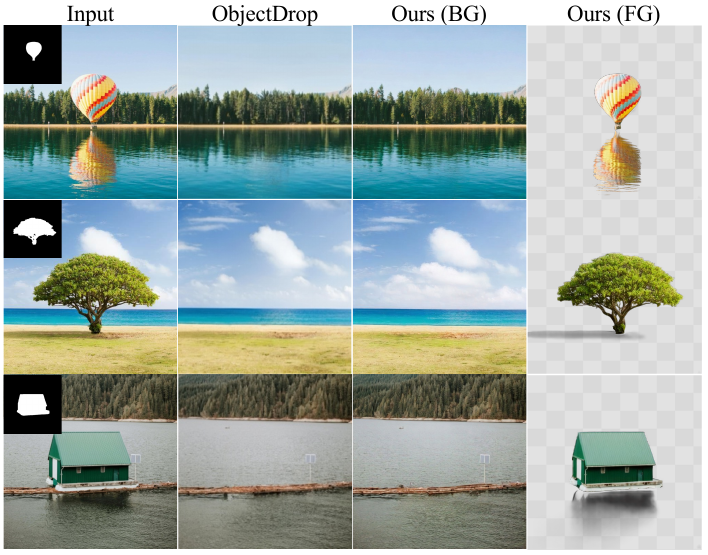}
   \caption{ \textbf{Object removal - comparison with ObjectDrop~\cite{liu2024object}.} Based on their released examples, our model demonstrates comparable quality in photorealistic object removal in the background layer, while decomposing the foreground with intact visual effects.}
   \label{fig:comp_object_removal_od}
\end{figure}

\subsection{Comparison on Object Removal}
To assess the quality of the background layers predicted by \ours{}, we evaluate the model on the object removal task, comparing it to several state-of-the-art approaches, including mask-based methods (ControlNet Inpainting~\cite{zhang2023adding}, SD-XL Inpainting~\cite{rombach2022high},  PowerPaint~\cite{zhuang2023task}, ObjectDrop~\cite{winter2024objectdrop}) and instruction-driven models (Emu-Edit~\cite{sheynin2024emu}, MGIE~\cite{fu2024mgie}, and OmniGen~\cite{xiao2024omnigen}). 

Quantitative evaluation among mask-based inpainting methods is conducted on three public benchmarks: \textbf{RORD}~\cite{sagong2022rord}, a real-world object removal dataset collected from video data with human-labeled loose and tight object masks; \textbf{MULAN}~\cite{tudosiu2024mulan}, a synthesized multi-layer dataset that provides instance-wise RGBA decompositions for COCO and LAION images; and \textbf{DESOBAv2}~\cite{liu2023desobav2}, a real-world image dataset with shadow mask annotations and synthesized image pairs where instance shadows are removed. Standard metrics, including PSNR, LPIPS, FID, and CLIP-FID, are used on RORD and MULAN, while regional similarity such as masked PSNR and masked SSIM are used for shadow removal in DESOBAv2. More details of dataset preparation will be included in the supplementary materials.

Most existing mask-based object removal approaches cannot automatically detect and remove visual effects associated with the target object, necessitating a loose mask input. Indeed, as shown in Tab.~\ref{tab:quant:remove}, using a loose mask introduces more inpainting area and thus hurts PSNR and LPIPS, but significantly improves result fidelity (i.e., FID and CLIP-FID) for all compared methods. In contrast, \ours{} demonstrates robustness to mask tightness and outperforms all methods by a large margin. For shadow removal on DESOBAv2, \ours{} surpasses other methods even without utilizing the shadow annotation included in the loose mask. As shown in Fig.~\ref{fig:comp_object_removal_mask_methods}, \ours{} provides substantial improvements over existing methods, generating more photorealistic background layers with fewer artifacts and minimal visual effect residues. Comparing with ObjectDrop, a leading model in object removal that can automatically eliminate visual effect with a tight input mask, we applied \ours{} to the example images released by their work. As shown in Fig.~\ref{fig:comp_object_removal_od}, \ours{} achieves similar high-quality backgrounds with effective object removal. Additionally, \ours{} decomposes the foreground with realistic visual effects, enabling further editing capabilities not supported by ObjectDrop. More visual comparisons will be included in the supplementary materials.

\begin{figure*}[t]
  \centering
   \includegraphics[width=0.92\linewidth]{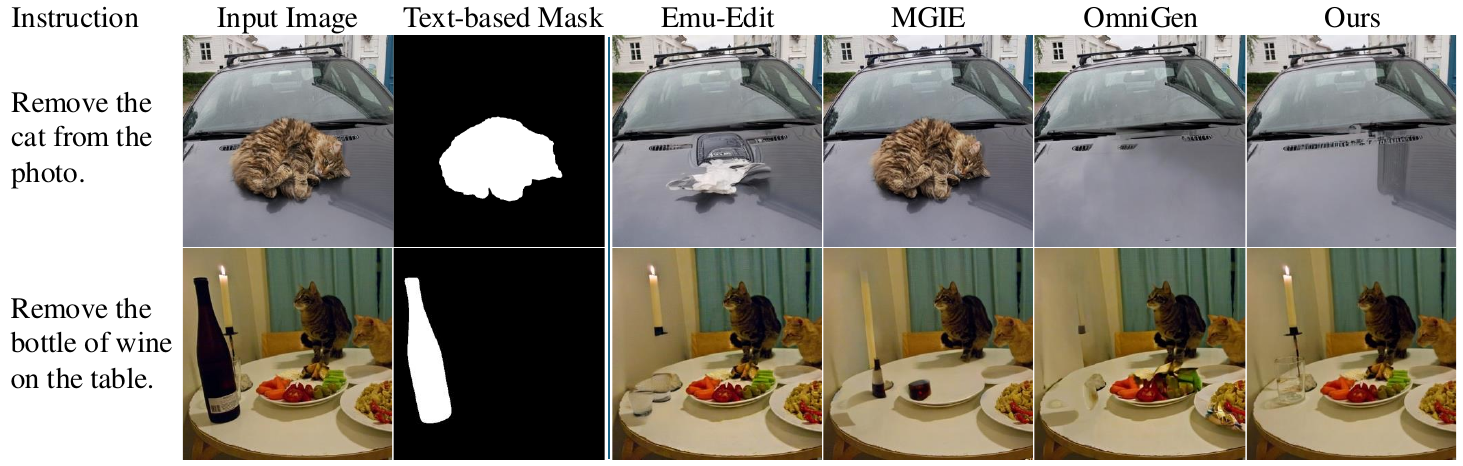}
   \caption{\textbf{Object removal - comparison with instruction-driven methods.} Combining with a text-based grounding method, our model can effectively remove target objects and preserve background integrity, while existing instruction-based editing methods, such as Emu-Edit~\cite{sheynin2024emu}, MGIE~\cite{fu2024mgie}, and OmniGen~\cite{xiao2024omnigen}, may struggle to fully remove the target or maintain background consistency.}
   \label{fig:com_object_removal_instructional_methods}
\end{figure*}

\begin{table}[t]
    \centering
    \setlength\tabcolsep{3pt}
    \renewcommand{\arraystretch}{1.0}
    \footnotesize
    \caption{\textbf{User study for instruction-driven object removal.} }
    \begin{tabular}{lcccc}
        \toprule
        \textbf{Methods} & \shortstack{\textit{Removal} \\ \textit{Effectiveness}} & \shortstack{\textit{Result} \\ \textit{Plausibility}} & \shortstack{\textit{Background} \\ \textit{Integrity}} & \shortstack{\textit{Overall}} \\
        \midrule
        Emu-Edit~\cite{sheynin2024emu} & $5.00\%$ & $4.38\%$ & $3.33\%$ & $4.79\%$ \\
        Ours & $57.08\%$ & $77.92\%$ & $76.25\%$ & $83.54\%$  \\
        \midrule
        OmniGen~\cite{xiao2024omnigen} & $4.07\%$ & $3.15\%$ & $2.96\%$ & $3.89\%$ \\
        Ours & $67.04\%$ & $80.56\%$ & $84.63\%$ & $87.78\%$ \\
        \bottomrule
    \end{tabular}
    \label{tab:user:remove}
\end{table}

\begin{figure*}[t]
  \centering
   \includegraphics[width=1.0\linewidth]{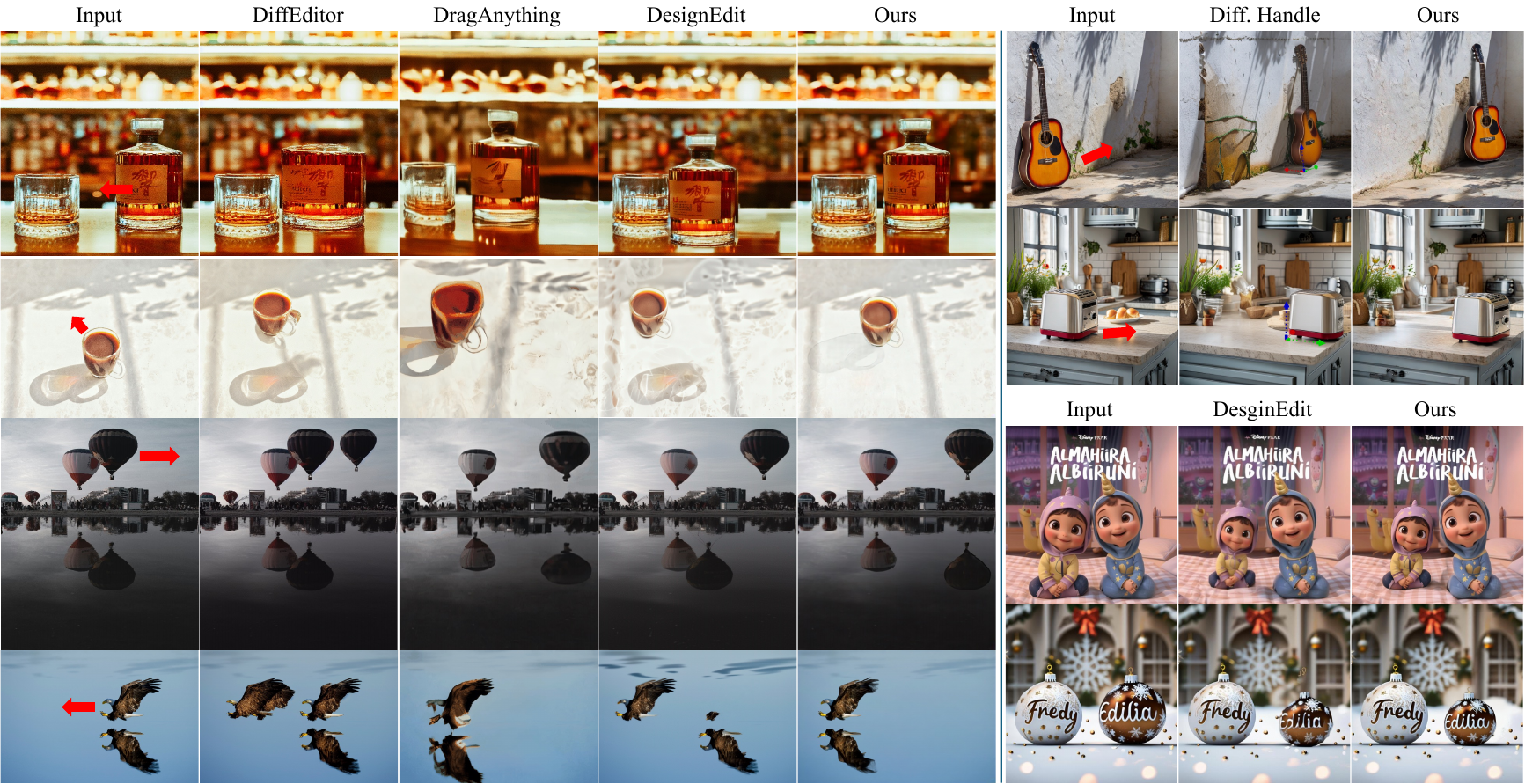}
   \caption{\textbf{Object spatial editing.} Our model enables precise object moving and resizing with seamless handling of visual effects, resulting in highly effective and realistic edits that preserve content identity. When applied to examples released by specific works, such as DiffusionHandle~\cite{pandey2024diffusion} and DesignEdit~\cite{jia2024designedit}, our model also achieves satisfying results.  }
   \vspace{-1em}\label{fig:object_spatital_editing}
   
\end{figure*}

\begin{figure*}[t]
  \centering
   \includegraphics[width=1.0\linewidth]{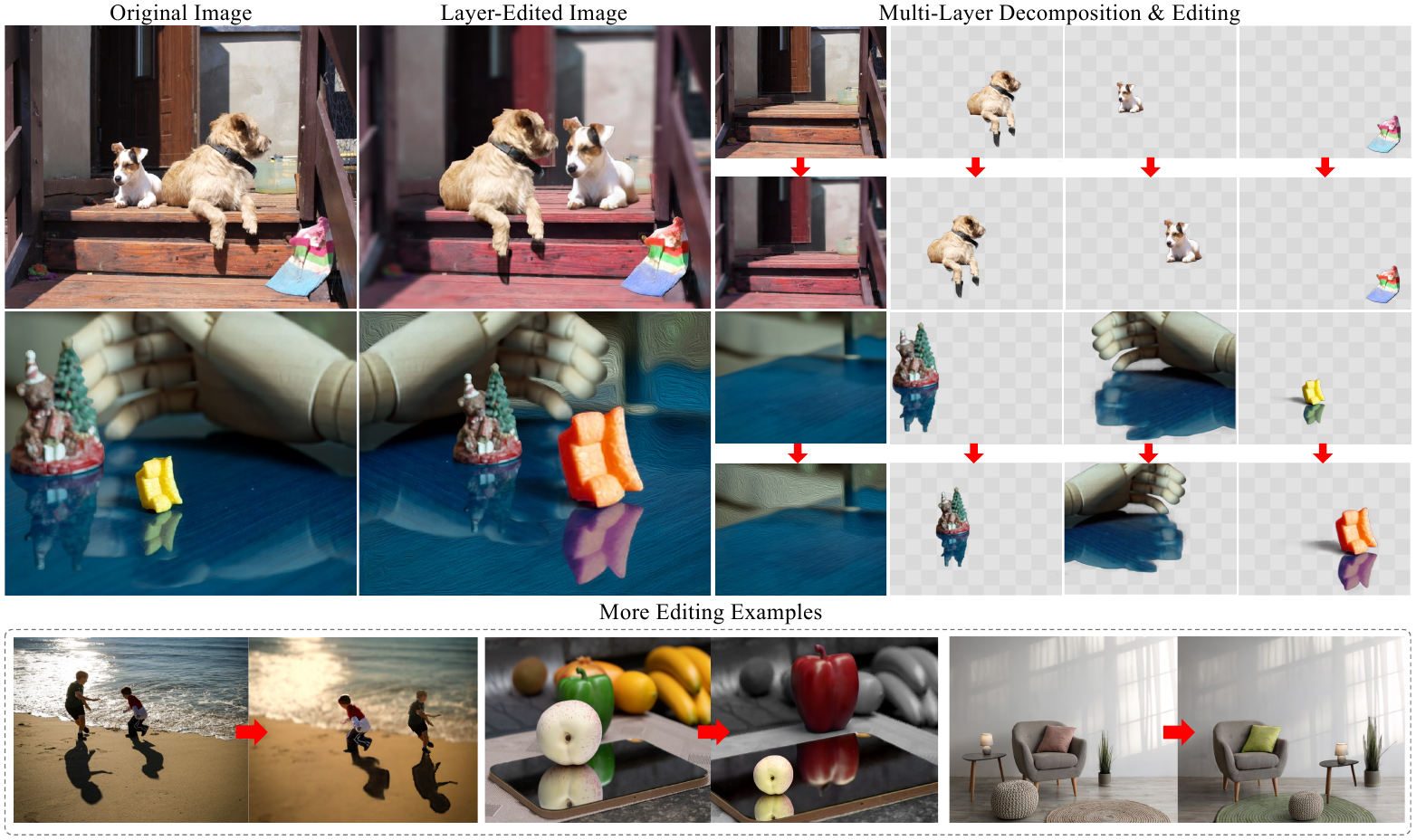}
   \caption{\textbf{Multi-layer Decomposition and Creative layer-editing.} By sequentially applying our model, we can decompose multiple foreground layers with distinct visual effects, which can then be used for further creative editing tasks.}
  
   \label{fig:creative_editing}
\end{figure*}

To compare with instruction-driven methods, we conduct a user study on $60$ randomly selected images from the \textbf{Emu-Edit Remove Set}~\cite{sheynin2024emu}. We ask $17$ independent researchers to compare results from \ours{} and an existing method, focusing on three quality aspects: \textit{removal effectiveness}, \textit{result plausibility}, and \textit{background integrity}. In cases where both methods achieve satisfactory results, users could mark it as a ``tie''. For a fair comparison, we use a grounding model to generate text-based masks to input to \ours{}. In total, $3060$ data points are collected in this study. As shown in Tab.~\ref{tab:user:remove}, \ours{} is clearly preferred in at least $83\%$ of the testing cases for overall quality. Example object removal results are visualized in Fig.~\ref{fig:com_object_removal_instructional_methods}. While Emu-Edit and MGIE struggle to fully remove the target object, OmniGen is more effective but does not reliably preserve the background integrity. Our model, in contrast, successfully removes the target object while preserving most background fine details.

\subsection{Comparison on Object Spatial Editing}

We further compare \ours{} for object spatial editing tasks, including object moving and resizing, against several state-of-the-art methods: DiffEditor~\cite{mou2024diffeditor}, DragAnything~\cite{wu2024draganything}, DesignEdit~\cite{jia2024designedit}, and Diffusion Handles~\cite{pandey2024diffusion}. For user study, we select $15$ editing samples using web images for each task, and ask $23$ independent researcher to compare results from \ours{} and an existing method, focusing on three quality aspects: \textit{edit effectiveness}, \textit{result plausibility}, and \textit{content integrity}. In this study, $2070$ data points are collected. As shown in Tab.~\ref{tab:user:edit}, \ours{} is at least $87\%$ more preferred in the spatial editing tasks, featuring superior plausibility in editing results. As shown in Fig.~\ref{fig:object_spatital_editing}, \ours{} enables seamless spatial editing for objects in various scenes. By preserving intact visual effects in the transparent foreground layer, shadows and reflections move naturally with the editing targets, allowing harmonious re-composition to be effortlessly achieved through alpha blending. When compared to DiffusionHandle~\cite{pandey2024diffusion} and DesignEdit~\cite{jia2024designedit} on their released examples, \ours{} demonstrates comparable results in most scenarios, including graphic design examples without requiring additional fine-tuning.

\begin{table}[t]
    \centering
    \setlength\tabcolsep{3pt}
    \renewcommand{\arraystretch}{1.0}
    \footnotesize
    \caption{\textbf{User study for object spatial editing.}}
    \begin{tabular}{llcccc}
        \toprule
        & \textbf{Methods} & \shortstack{\textit{Edit} \\ \textit{Effectiveness}} & \shortstack{\textit{Result} \\ \textit{Plausibility}} & \shortstack{\textit{Content} \\ \textit{Integrity}} & \shortstack{\textit{Overall}} \\
        \midrule
        \multirow{4}{*}{\rotatebox[origin=c]{90}{\textbf{Moving}}} & DesignEdit~\cite{jia2024designedit} & $1.67\%$ & $1.11\%$ & $1.11\%$ & $2.22\%$ \\
        & Ours & $71.67\%$ & $90.56\%$ & $77.22\%$ & $94.44\%$ \\
        \cmidrule(lr){2-6}
        & DragAnything~\cite{wu2024draganything} & $1.82\%$ & $1.21\%$ & $2.42\%$ & $1.21\%$\\
        & Ours & $67.88\%$ & $93.33\%$ & $95.15\%$ & $96.36\%$ \\
        \midrule
        \multirow{4}{*}{\rotatebox[origin=c]{90}{\textbf{Resizing}}} & DesignEdit~\cite{jia2024designedit} & $2.22\%$ & $3.89\%$ & $3.33\%$ & $3.89\%$ \\
        & Ours & $69.44\%$ & $82.78\%$ & $71.67\%$ & $87.22\%$\\
        \cmidrule(lr){2-6}
        & DiffEditor~\cite{mou2024diffeditor} & $1.11\%$ & $1.11\%$ & $1.21\%$ & $1.11\%$ \\
        & Ours & $95.15\%$ & $96.36\%$ & $92.21\%$ & $96.36\%$ \\
        \bottomrule
    \end{tabular}
    \label{tab:user:edit}
\end{table}

\subsection{Multi-Layer Decomposition and Editing}
 \ours{} can be applied sequentially to an original input image with different instance masks, decomposing multiple layers along with their visual effects, as examples shown in Fig.~\ref{fig:creative_editing}. 
This process enables creative and complex layer-based editing for each individual layer, including spatial manipulation, recoloring, and filtering. Once editing is complete, the re-composition of all layers maintains a natural and realistic appearance, as demonstrated in Fig.~\ref{fig:creative_editing}.

%% file: sec/5_conclusion.tex
\section{Discussions}
\label{sec:conclusion}

In conclusion, our model achieves high-quality image layer decomposition and outperforms existing state-of-the-art methods in object removal and spatial editing across multiple benchmarks and user studies. By decomposing images into a photorealistic background an a transparent foreground with faithfully preserved visual effects, our model unlocks various creative possibilities for layer-wise image editing. 
The proposed consistency loss enables effective learning of accurate representations for the transparent foreground layer with visual effects, even in the absence of ground-truth data which is challenging to collect from real-world images. While our current dataset preparation pipeline focuses primarily on common visual effects, such as shadows and reflections, extending this work to include other effects, such as smoke and mist, remains an exciting avenue for future exploration.

%% file: sec/X_suppl.tex
\clearpage
\setcounter{page}{1}
\maketitlesupplementary

\begin{figure*}[t]
  \centering
   \includegraphics[width=1.0\linewidth]{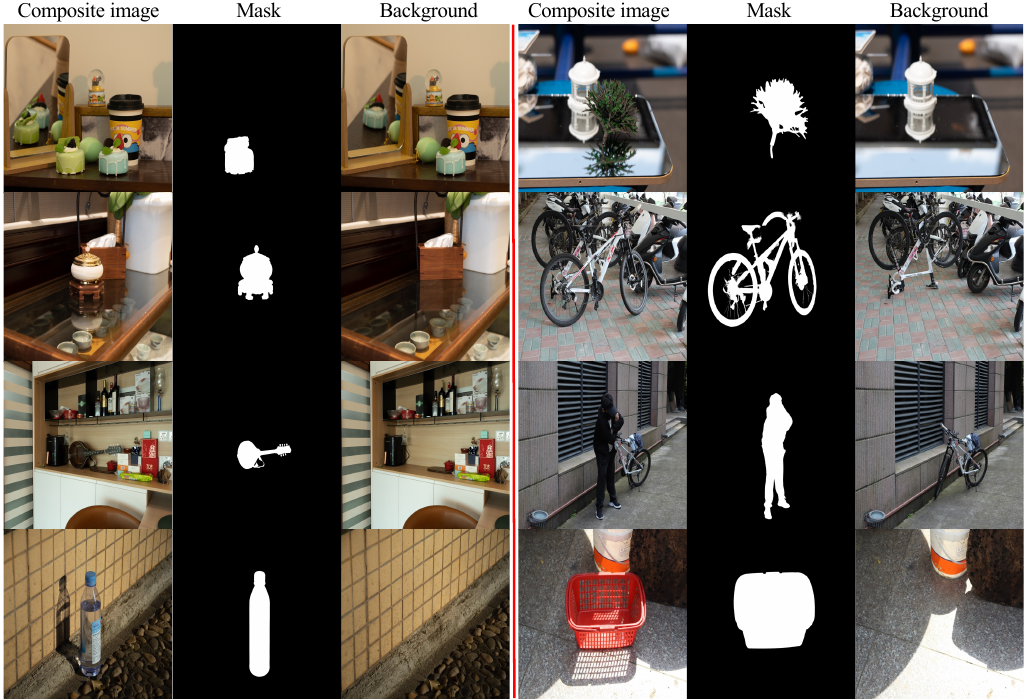}
   \caption{Examples of the real-world camera-captured image pairs used in our model training and ablation study. Images are captured by a camera in real-world scenes. The masks are manually annotated, indicating the target objects to remove. }
   \vspace{-1em}\label{fig:held_out_testset}
   
\end{figure*}

\begin{table}[t]
    \setlength\tabcolsep{2pt}
    \renewcommand{\arraystretch}{1.0}
    \small
    \centering
    \caption{Additional ablation study of \ours{} on held-out test set based on random re-composition.}
    \begin{tabular}{lcc}
        \toprule
        Model & FID $\downarrow$ & CLIP-FID $\downarrow$ \\
        \midrule
        $V_0$:RGB-only & - & - \\
        $V_1$:$V_0+$RGBA FG (obj.) & $45.758$ & $3.756$ \\
        $V_2$:$V_0+$RGBA FG (obj.+v.e.) & $45.123$ & $3.739$ \\
          \midrule
     Ours:$V_2+\mathcal{L}_{\text{consist}}$ & $44.260$ & $3.173$ \\
        \bottomrule
    \end{tabular}
    \label{tab:rand_composite_ablations}
\end{table}

\begin{figure*}[t]
  \centering
   \includegraphics[width=1.0\linewidth]{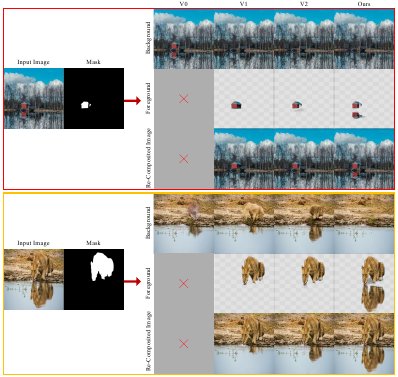}
   \caption{Visualization of results generated by model variants presented in Table~\ref{tab:quant:ablation_study} of the main manuscript. Our model can generate higher-quality foreground and background layers and produce more plausible and realistic re-composite results by effectively modeling the visual effects.}
   \vspace{-1em}\label{fig:ablation_vis}
   
\end{figure*}

\section{Additional Results for the Ablation Study}
\label{sec:additional_ablation}
\textbf{Test Set Details.} The test set is a held-out subset of our camera-captured data consisting of 635 image pairs (composite image and background image). To construct this dataset, we manually collected real-world examples comprising photos of scenes captured before and after the removal of an object, while ensuring all other elements in the scene remain unchanged. We also manually labeled the binary object mask for the removal target. As illustrated in Fig.~\ref{fig:held_out_testset}, the dataset encompasses both indoor and outdoor scenarios, effectively reflecting real-world phenomena such as shadows and reflections. This test set allows us to evaluate not only the quality of the decomposed background naturally but also the quality of the foreground. By re-compositing the background and foreground layer output, we can effectively assess the fidelity and visual coherence of the foreground, including the visual effect components.

\textbf{Qualitative Comparison.} To more intuitively demonstrate the effectiveness of our design in \ours{}, in addition to the quantitative analysis represented in Table~\ref{tab:quant:ablation_study} of the main manuscript, we provide more visual results of the four model variants in Fig.~\ref{fig:ablation_vis}. For each variant, we present the decomposed background and foreground layers, along with the re-composited image obtained by alpha blending the two layers. For the RGB-only model ($V_0$), which lacks an RGBA foreground, we show only the decomposed background for reference. From the visual comparison among $V_1$, $V_2$, and ``Ours'', it is evident that our method, which leverages consistency loss to explicitly model visual effects in the foreground layer, produces: (i) background layers with cleaner removal and less artifacts, (ii) foreground layers with more accurate extraction of transparent visual effects, resulting in re-composited results that are more plausible and realistic.

\textbf{Quantitative Comparison}. To more comprehensively evaluate the quality of the decomposed foreground, we randomly move/resize the foreground prediction and then re-composite it onto the decomposed background to evaluate the fidelity of the resulting image. Specifically, there are three parameters to randomly adjust: $\Delta{X}$, $\Delta{Y}$, and $\Delta{S}$. $\Delta{X} \in [-0.3,+0.3]$ and $\Delta{Y} \in [-0.3,+0.3]$ specify the horizontal and vertical location changes as proportions of the input dimensions, while $\Delta{S} \in [0.5,1.5]$ specifies a scaling ratio w.r.t. the original size. For each image, we randomly select three parameters and apply the same adjustment to all model variants' foreground prediction. The FID and CLIP-FID of the randomly re-composite images are reported in Table~\ref{tab:rand_composite_ablations}. Comparing with other model variants, leveraging consistency loss to explicitly model visual effects in the foreground layer indeed improves re-composition quality.

\begin{table}[t]
    \setlength\tabcolsep{2pt}
    \renewcommand{\arraystretch}{1.0}
    \small
    \centering
    \caption{Comparison of \ours{} with instruction-driven object removal methods on Emu-Edit Remove Set~\cite{sheynin2024emu}.}
    \begin{tabular}{lcc}
        \toprule
        Model & FID $\downarrow$ & CLIP-FID  $\downarrow$ \\
        \midrule
        Emu-Edit~\cite{sheynin2024emu} & $47.555$ & $6.711$ \\
        OmniGen~\cite{xiao2024omnigen} & $48.116$ & $6.283$ \\
        Ours& $38.998$ & $5.622$ \\
        \bottomrule
    \end{tabular}
    \label{tab:quan_instructional_models}
\end{table}

\section{Additional Results for the Mask-Based Object Removal Experiment}
\label{addition_vis_res_benchmarks}

\textbf{Benchmarks Details.} Here, we provide more details for the mask-based object removal benchmarks used to calculate the metrics presented in Table~\ref{tab:quant:remove} of the main manuscript. 

\begin{figure*}[t]
  \centering
   \includegraphics[width=0.65\linewidth]{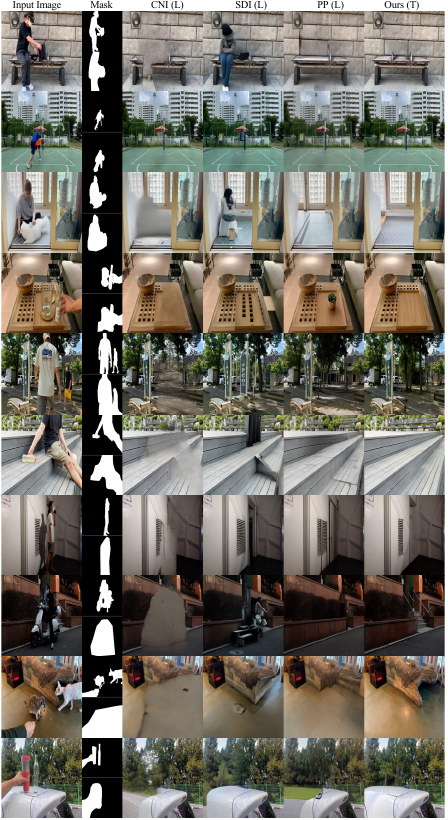}
   \caption{ \textbf{Object removal - comparison with mask-based methods on the RORD dataset.} Our model, using tight input masks, generates more visually plausible results with fewer artifacts compared to ControlNet Inpainting~\cite{zhang2023adding}, SD-XL Inpainting~\cite{rombach2022high}, and PowerPaint~\cite{zhuang2023task}, which all require loose mask input. }
   \vspace{-1em}\label{fig:rord_removal}
   
\end{figure*}

\begin{figure*}[t]
  \centering
   \includegraphics[width=0.65\linewidth]{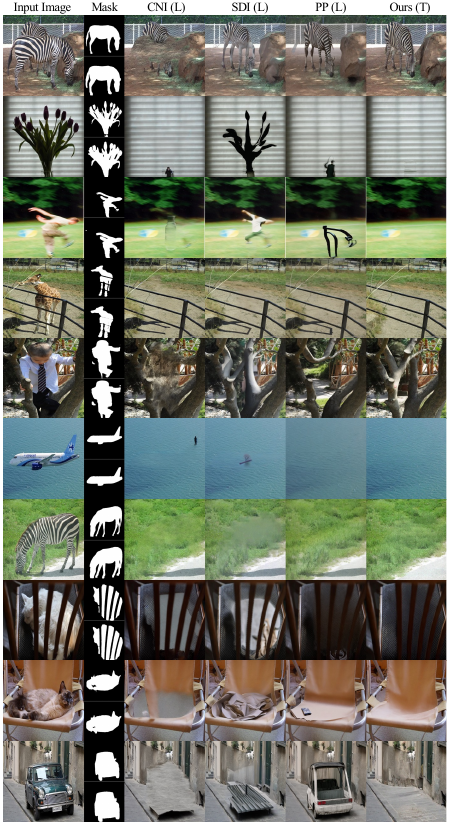}
   \caption{ \textbf{Object removal - comparison with mask-based methods on the MULAN dataset.} Our model, using tight input masks, generates more visually plausible results with fewer artifacts compared to ControlNet Inpainting~\cite{zhang2023adding}, SD-XL Inpainting~\cite{rombach2022high}, and PowerPaint~\cite{zhuang2023task}, which all require loose mask input.}
   \vspace{-1em}\label{fig:mulan_set_2_removal}
   
\end{figure*}

\begin{figure*}[t]
  \centering
   \includegraphics[width=0.68\linewidth]{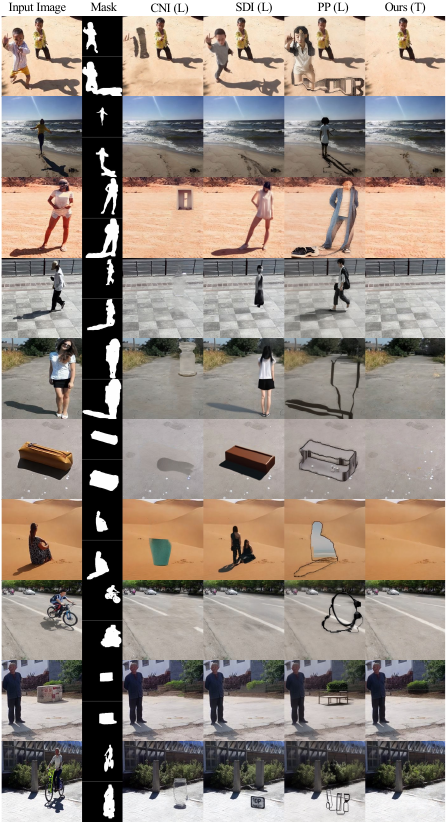}
   \caption{  \textbf{Object removal - comparison with mask-based methods on the DESOBAv2 dataset.} Our model, using tight input masks, generates more visually plausible results with fewer artifacts compared to ControlNet Inpainting~\cite{zhang2023adding}, SD-XL Inpainting~\cite{rombach2022high}, and PowerPaint~\cite{zhuang2023task}, which all require loose mask input.}
   \vspace{-1em}\label{fig:desobav2_removal}
   
\end{figure*}

\begin{itemize}
    \item RORD~\cite{sagong2022rord}: We randomly select 1,029 images from the original test set to reduce data redundancy caused by sampling from the same video. The dataset provides both manually labeled loose masks and tight masks for the real-world object removal task.  The average area of the loose mask is 3.70 times that of the tight mask in each image. As shown in Fig.~\ref{fig:rord_removal}, RORD includes diverse indoor and outdoor scenes, featuring removal of various target objects with soft shadows or reflections in real-world settings.
    \item MULAN~\cite{tudosiu2024mulan}: We randomly select 1,000 images from MULAN-COCO for our evaluation. For each image, the dataset provides multiple object layers in RGBA format, and we select the object in the top most layer as the removal target. To reduce hallucination problems in traditional inpainting methods caused by tight object mask, we further dilate the object mask by 10 pixels. As shown in Fig.~\ref{fig:mulan_set_2_removal}, MULAN data also includes diverse indoor and outdoor scenes, featuring object removal in more cluttered settings.
    \item DESOBAv2~\cite{liu2023desobav2}: There are 750 images in the test set including binary object masks and paired shadow masks. We use the binary object masks as tight mask to input to \ours{} and merge the object mask and the corresponding shadow mask to create loose mask to input to other inpainting methods. Similarly, to reduce hallucination problems in traditional inpainting methods, the loose masks are further dilated by 10 pixels. The average area of the loose mask is 2.35 times that of the tight mask. As shown in Fig.~\ref{fig:desobav2_removal}, DESOBAv2 mostly features outdoor scenes with hard object shadows cast on surfaces with different materials and textures, adding more challenges to decompositing the visual effects.
\end{itemize}

\textbf{More Visual Results.}
More visual comparison with ControlNet Inpainting~\cite{zhang2023adding}, SD-XL Inpainting~\cite{rombach2022high}, and PowerPaint~\cite{zhuang2023task} on the three public benchmarks is provided in Fig.~\ref{fig:rord_removal}, Fig.~\ref{fig:mulan_set_2_removal}, and Fig.~\ref{fig:desobav2_removal}. It can be observed that, with the assistance of the loose mask, the three baselines are able to remove most parts of the target object. However, they struggle to eliminate it entirely and face challenges in removing the shadows associated with the target object. Additionally, achieving photorealistic background completion in human plausible style remains a significant challenge. In contrast, our model, using only the tight mask, performs consistently better across a wide range of data sources.

\section{Additional Results for the Instruction-Driven Object Removal Experiment}
\begin{figure*}[t]
  \centering
   \includegraphics[width=1.0\linewidth]{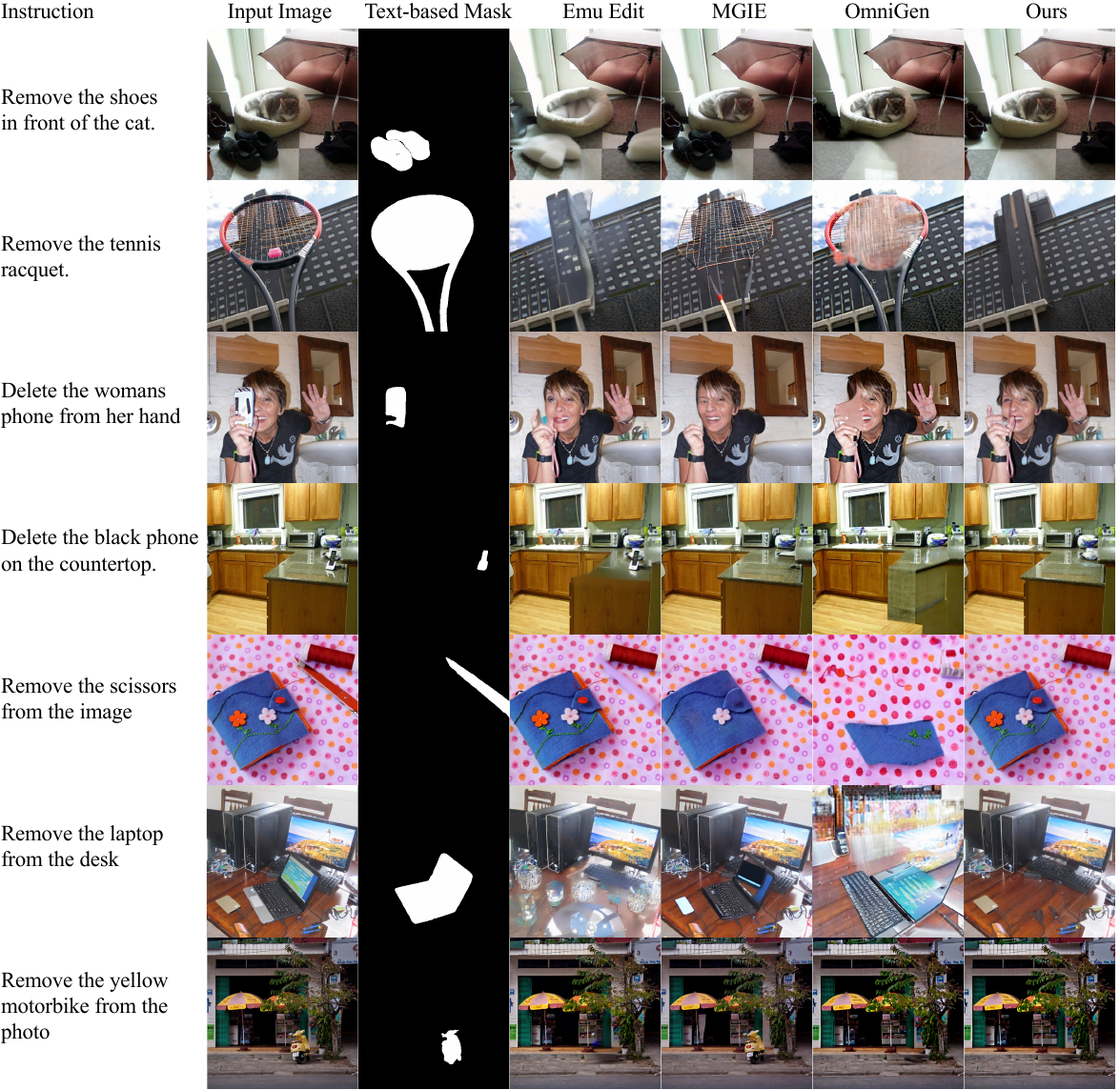}
   \caption{ \textbf{Object removal - more comparison with instruction-driven methods on Emu-Edit~\cite{sheynin2024emu} removal set.} Combining with a text-based grounding method, our model can effectively remove target objects and preserve background integrity, while existing instruction-based editing methods, such as Emu-Edit~\cite{sheynin2024emu}, MGIE~\cite{fu2024mgie}, and OmniGen~\cite{xiao2024omnigen}, may struggle to fully remove the target or maintain background consistency.}
   \vspace{-1em}\label{fig:instructional_removal}
   
\end{figure*}

\textbf{Qualitative Comparison.} Fig.~\ref{fig:instructional_removal} presents additional comparison results with instruction-driven methods on the object removal task on Emu-Edit Remove Set~\cite{sheynin2024emu,emu_edit_test_set_generations}. Beyond showcasing the superior object removal performance of our model, these results further highlight its enhanced background integrity and completion capabilities.

\textbf{Quantitative Comparison.} We also perform a quantitative comparison with instruction-driven methods, specifically Emu-Edit~\cite{sheynin2024emu} and OmniGen~\cite{xiao2024omnigen}. Using the released generation results from Emu-Edit Remove Set~\cite{emu_edit_test_set_generations}, we evaluate the performance based on FID and CLIP-FID metrics. For a fairer comparison, we use text-based masks as input to our model. As shown in Table~\ref{tab:quan_instructional_models}, our model outperforms existing approaches by a large margin.

\section{More Image Layer Decomposition Results from \ours{}}
\begin{figure*}[t]
  \centering
   \includegraphics[width=1.0\linewidth]{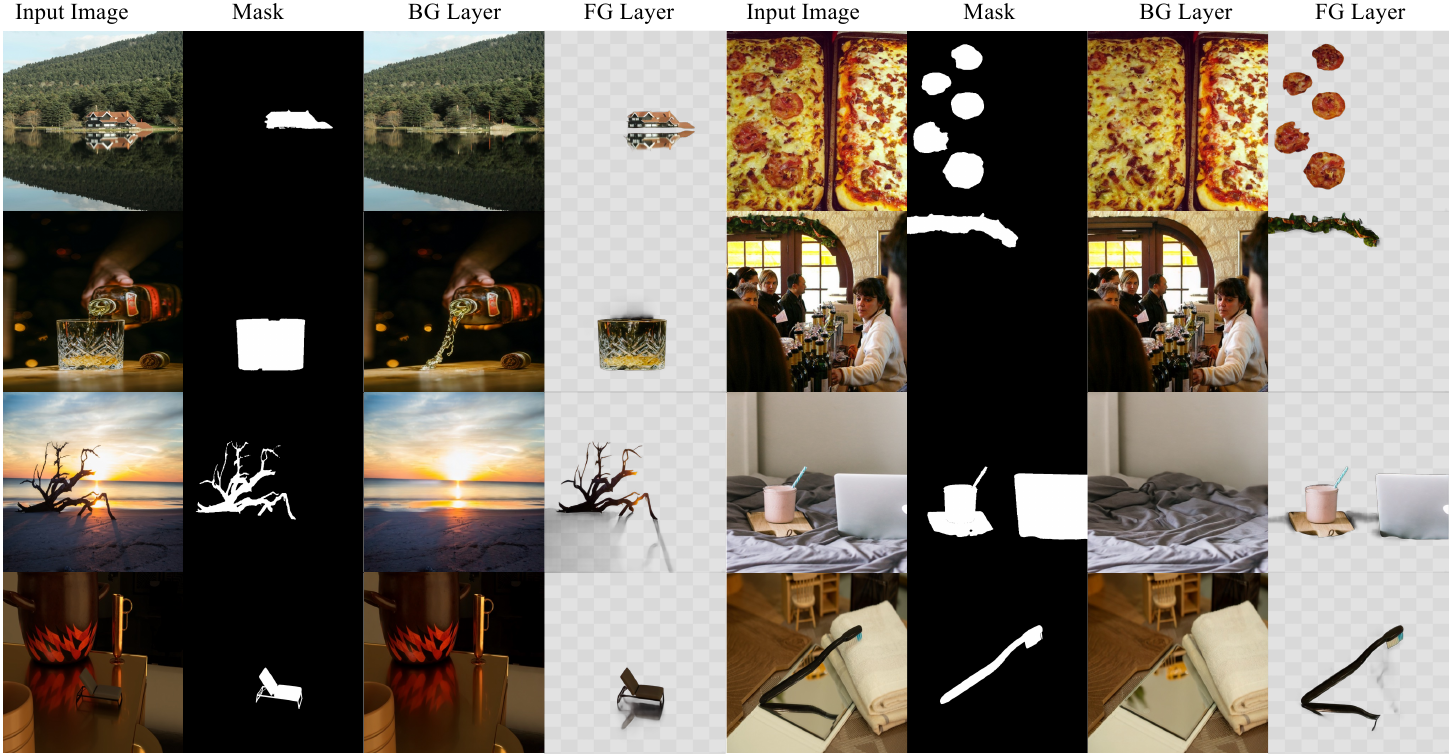}
   \caption{Additional image decomposition results of our model on public benchmarks and web images. These results demonstrate the robustness of our model across diverse data sources.}
   \vspace{-1em}\label{fig:our_removal_examples}
   
\end{figure*}

As shown in Fig.~\ref{fig:our_removal_examples}, we provide comprehensive visualization results from various data sources, including web images, public datasets, and the held-out test set. These results demonstrate that our model is robust across diverse scenarios.

